\documentclass[letterpaper, 10 pt, conference]{ieeeconf}  % Comment this line out if you need a4paper

\usepackage{graphicx}
\usepackage{bm}
\usepackage{amsmath,amsfonts}
\usepackage{amssymb}
\usepackage[caption=false,font=normalsize,labelfont=sf,textfont=sf]{subfig}
\usepackage{booktabs} 
\usepackage{caption}
\IEEEoverridecommandlockouts                              
\title{\LARGE \bf
Robotic Servo Tracking of Moving Targets with Dynamic Imitation Constraints
}

\author{Yazhe Luo$^{1,3}$, Sipu Ruan$^{1}$, Yifei Li$^{1}$ and Diansheng Chen$^{1,2}$ %   <-this % stops a space
\thanks{This work was supported by the National Key Research and Development Program of China (2024YFB4709800), National Natural Science Foundation of China (No. 52505001) and the Fundamental Research Funds for the Central Universities.(Corresponding author: Sipu Ruan and Diansheng Chen)}% <-this % stops a space
\thanks{$^{1}$Yazhe Luo, Sipu Ruan, Yifei Li and Diansheng Chen are with the Robotic Institute, School of Mechanical Engineering and Automation, Beihang University, Beijing 100191, China (e-mail: by2007124@buaa.edu.cn; ruansp@buaa.edu.cn; liyifei18810000840@buaa.edu.cn; chends@buaa.edu.cn)}%
\thanks{$^{2}$Diansheng Chen is also with the Hunan Intelligent Rehabilitation Robot and Auxiliary Equipment Engineering Technology Research Center, Changsha 410004, China}%
\thanks{$^{3}$Yazhe Luo is also with the College of Robotics Science and Engineering, Taiyuan University of Technology, Taiyuan City, Shanxi Province, 030024, China}%
}

\begin{document}

\maketitle
\thispagestyle{empty}
\pagestyle{empty}

%%%%%%%%%%%%%%%%%%%%%%%%%%%%%%%%%%%%%%%%%%%%%%%%%%%%%%%%%%%%%%%%%%%%%%%%%%%%%%%%
\begin{abstract}

Imposing explicit trajectory constraints in robot visual servoing remains challenging. Existing tracking methods achieve fast responses by mapping visual residuals to control velocities, but they have weak constraints on the intermediate motion process, which lead to trajectory discontinuity, oscillation, or conservative behaviors. To enable constrained tracking for moving targets, this paper proposes a servo tracking method based on imitation trajectory constraints. 
A dynamic model describing the robot approaching a moving target is formulated and analyzed for convergence. A time-scalable deformation mechanism and a trajectory modulation incorporating shape and amplitude components are introduced to generate a series of trajectories in real time, from which tracking points are adaptively determined to form dynamic constraints. The robot velocity is then computed from target pose differentials or tracked key features to follow the constrained trajectory.
Simulation and real-world experiments demonstrate that the proposed method can achieve dynamic obstacle avoidance and high-precision convergence compared with several state-of-the-art methods in complex environments.

\end{abstract}

%%%%%%%%%%%%%%%%%%%%%%%%%%%%%%%%%%%%%%%%%%%%%%%%%%%%%%%%%%%%%%%%%%%%%%%%%%%%%%%%
\section{INTRODUCTION}
Tracking moving targets is a fundamental yet challenging problem in robotic manipulation, which arises in application scenarios such as household services, logistics sorting, and human–robot collaboration. Compared with static operation, robotic tracking requires continuous perception of object's motion and real-time control strategies. Therefore visual servoing based on object's features has been adopted in robotic tracking due to its closed-loop nature and robustness \cite{ref_Comparison, ref_Image_Based, ref_Safety_Control, ref_Dynamic_Tracking}. Currently, existing tracking frameworks are built upon an instantaneous error-minimization mechanism, typically adopting image-based visual servoing methods, where the control velocity is directly driven by the residual between the initial state and the desired target state \cite{ref_stereo, ref_precise}. And the residual is defined by tracking texture points, pose coordinate or bounding box vertices of target between the desired feature ${{\bm{p}'}_{c}}$ and the current feature ${\bm{p}_{c}}$, as
\begin{equation}
\label{visual_servo_eq1}
\Delta {\bm{p}_{c}}={{\bm{p}'}_{c}}-{\bm{p}_{c}}.
\end{equation}

Then, the point's speed ${{\dot{\bm{p}}}_{c}}={\Delta {\bm{p}_{c}}}/{\Delta t}\;$ is determined according to the desired convergence time $\Delta t$. The relationship between velocities of $n$ feature points and object's spatial velocity ${\bm{V}_{o}}={{\left( {\bm{v}_{o}}^\text{T},{\bm{\omega }_{o}}^\text{T} \right)}^\text{T}}$, where $\bm{v}_{o}$ and $\bm{\omega}_{o}$ denote the linear and angular velocities, can be expressed as:
\begin{equation}
\label{visual_servo_eq2}
\begin{aligned}
 \left\{ \begin{aligned}
 {{\bm{\dot{p}}}_{c1}}={{J}_{c1}}{\bm{V}_{o}} \\ 
 {{\bm{\dot{p}}}_{c2}}={{J}_{c2}}{\bm{V}_{o}}  \\ 
 \cdot \cdot \cdot \text{ }\text{ }\text{ }\text{ }\text{ }\\ 
 {{\bm{\dot{p}}}_{cn}}={{J}_{cn}}{\bm{V}_{o}} \\ 
\end{aligned} \right.\to {{\bm{\dot{P}}}_{c}}=L{\bm{V}_{o}}, \\ 
 {{\bm{\dot{P}}}_{c}}\in {{\mathbb{R}}_{2n\times 1}}\text{, }L\in {{\mathbb{R}}_{2n\times 6}}\text{, }{\bm{V}_{o}}\in {{\mathbb{R}}_{6\times 1}}, \\ 
\end{aligned}
\end{equation}
where, $L$ is composed of multiple image Jacobian matrices ${{J}_{ci}}$, which constitutes the core of the visual-servo control scheme, and $\bm{\dot{P}}$ consists of multiple $\bm{\dot{p}}_{ci}$. Subsequently, multiple feature points are employed to construct the linear non-homogeneous overdetermined system, from which $\bm{v}$ is obtained via an optimization-based solution.
For example, in the recent data-driven paradigm, Adrian et al. \cite{ref_ibvs_adrian} employed a self-supervised shared encoder to extract keypoints and achieved tracking of moving objects through velocity mapping. In addition, Yu \cite{ref_yu} proposed a hypernetwork-based neural controller that directly outputs servo velocities, enabling multi-pose tracking. Compared with traditional open-loop control, such direct or weakly constrained velocity-mapping approaches achieve higher accuracy and faster response. However, they mainly consider the initial point of trajectory and the endpoint of moving target, while imposing limited explicit constraints on the overall trajectory.
For example, in tasks where the robot grasps while moving, relative motion exists between robot and target. In this case, grasping target in a cluttered scene requires both real-time tracking and obstacle avoidance.
In addition, the residual fluctuation leads to oscillation, discontinuity, or overly conservative behavior \cite{ref_visual}. Therefore, we aim to further enhance the capability of constraining the overall dynamic trajectory.

Recent studies have introduced trajectory-level constraints into visual servoing frameworks. One line of research leverages optimization-based or predictive control techniques to encode velocity, acceleration, visibility, and safety constraints into the control law, thereby improving local motion smoothness and safety \cite{ref_shao, ref_Uncalibrated, ref_Learning_demonstration}. For example, Shao \cite{ref_shao} utilizes target’s prior velocity to construct a cost function for solving local motion state, and achieves high-precision tracking with natural motion characteristics. However, motion of object contains numerous non-differentiable points, which lead to deviations when computing cost function. Thus, such approaches exhibit limited adaptability to motion of object and are mainly applicable to scenarios involving conveyor belts or predefined motion patterns.
 
In parallel, another line of work adopts sampling-based planning or image-based roadmap methods to generate trajectories that satisfy visibility and collision constraints in advance or online \cite{ref_adaptive_framework, ref_neural_dynamics, ref_lyapunov_based}. 
Within this category, some approaches formulate trajectory as a global optimization problem, where motion states are jointly optimized to ensure smoothness and safety. Representative examples include OMPL-based sampling planners, where CHOMP and STOMP are employed as post-processing optimization \cite{ref_OMPL}. They optimize the entire trajectory from the initial to the final point. Although they are capable of handling moving targets, global optimization leads to computational burden and limited responsiveness. Such approaches struggle to adapt to rapid or irregular target motion, and accuracy is constrained to the centimeter level. 
Meanwhile, learning-based methods, particularly imitation learning and reinforcement learning, have been explored in visual servoing. Some learn stable tracking systems from demonstrations \cite{ref_Constrained, ref_roadmaps, ref_safe_smooth}, while others employ reinforcement learning to map visual observations to velocity commands, exhibiting strong adaptability\cite{ref_safety_critical, ref_hannes_imitation}. 
From trajectory constraint perspective, imitation learning can preserve motion smoothness while generalizing trajectories toward different endpoints, such as trajectory-primitive-based fitting methods \cite{ref_Gauss_series, ref_GMM}. 
Nevertheless, these trajectory constraints are incorporated implicitly through reward design or post-processing modules. Explicit modeling of trajectory shape, temporal consistency, and continuity during online replanning remains limited, making it difficult to satisfy the stability and interpretability requirements of tracking \cite{ref_roadmaps, ref_servo_predictive}. 

From the above analysis, it can be concluded that robotic tracking requires not only fast responses, but also explicit motion trajectory constraints. To this end, this paper focuses on the following aspects:

\begin{itemize}

\item We formulate the end-effector tracking toward a moving target as a time-varying system, and provide a convergence analysis that guarantees stable regulation under bounded target motion.

\item We propose a dynamic constrained trajectory (DCT) that integrates temporal scaling with shape–amplitude decoupled modulation and enforces a monotonic progress property, enabling real-time adaptation to moving targets while preserving motion continuity for smooth, non-jerky tracking.

\item We design a constraint-aware, multi-feature servo controller that tracks the DCT, and we validate it through both simulation and real-world tracking experiments, demonstrating reduced oscillation and overshoot as well as higher success rates over baselines.

\end{itemize}

\begin{figure}[!t]
\centering
\includegraphics[width=3.4in]{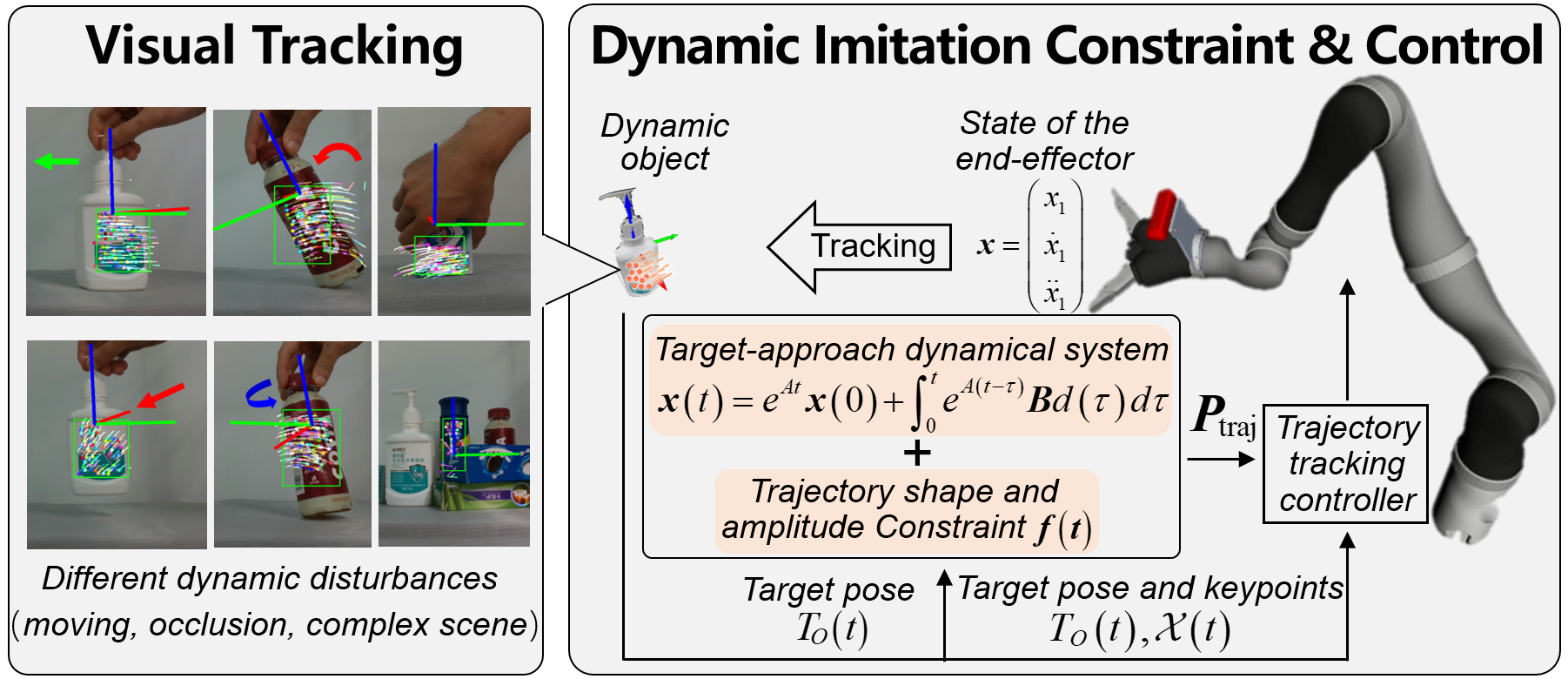}
\caption{The framework of constraint-tracking control method}
\label{total}
\end{figure}

\section{PRELIMINARIES}

Since \eqref{visual_servo_eq1} and \eqref{visual_servo_eq2} map 2D feature points to robot control velocities without explicit constraint formulation, it is necessary to reformulate the tracking problem from a system dynamics perspective. To achieve refined control of dynamic trajectories, we represent the process of the end-effector approaching moving target by a state-space equation. For the six motion dimensions of the manipulator end-effector (three translational and three rotational Euler angles), we first model one representative dimension, while the remaining dimensions follow the same formulation. Accordingly, the state of tracking process can be expressed as
\begin{equation}
\label{state_eq}
\dot{\bm{x}}=A\bm{x}\left( t \right)+\bm{B}d(t),
\end{equation}
\begin{equation}
\label{}
A=\!\left( \begin{matrix}
   0 & 1  \\
   -\frac{k}{m} & -\frac{c}{m}  \\
\end{matrix} \right),\bm{B}\!=\!\left( \begin{matrix}
   0  \\
   \frac{k}{m}  \\
\end{matrix} \right),\bm{x}\!\left( t \right)\!=\!\left( \begin{matrix}
   {{x}_{1}}  \\
   {\dot{x}_{1}}  \\
\end{matrix} \right),
\end{equation}
where:
\begin{itemize}
\item $A$ is the system matrix;
\item ${d}(t)$ is the input associated with the target motion (one corresponding motion dimension selected);

\item ${x}_{1}$, $\dot{x}_{1}$ and ${\ddot{x}_{1}}$ denote the position, velocity, and acceleration of the selected motion dimension of end-effector, respectively;

\item $k$, $c$, and $m$ represent the positive real constants (thus $A$ is invertible).

\end{itemize}

Under these definitions, the solution of system is given by 
\begin{equation}
\label{solve_state_eq}
{\bm{x}}\left( t \right)={{e}^{At}}{\bm{x}}\left( 0 \right)+\int_{0}^{t}{{{e}^{A\left( t-\tau  \right)}}}{\bm{B}d}\left( \tau  \right)d\tau. 
\end{equation}

Since instantaneous error minimization alone is insufficient to regulate trajectory and ${d}(t)$ lacks an explicit analytical expression. Therefore, the problem addressed in this paper is formulated as follows. 

\begin{itemize}
\item 
Construct a dynamic constrained trajectory (DCT) while preserving continuity and ensuring convergent tracking toward the moving target;

\item
Introduce dynamic trajectory constraint to handle task-specific requirements(e.g., obstacle avoidance during human-like tracking);

\item
Compute a constraint-aware control velocity to ensure convergent tracking.

\end{itemize}

\section{APPROACH}
Before introducing the proposed framework, we clarify the perception prerequisite of this study. Feature points, pose and bounding boxes of target can be extracted by traditional detection methods or deep-learning-based approaches for guiding and constraining robotic motion \cite{robot_detection}. The framework of constraint-tracking control comprises the following key modules as shown in Fig.~\ref{total}: a convergence model for approaching moving target, imitation trajectory constraints introduced in tracking, and motion controller for constraints.

\subsection{The Convergence of Dynamic Model}

As indicated, the system is influenced not only by the intrinsic parameters but also by the input motion of target. 
Although ${d}(t)$ is uncertain during the intermediate stage and does not admit an explicit analytical expression, successful execution requires stable convergence at the final state. Therefore, we assume that there exists final stable state ${d}_{f}$ such that 
${\mathop{\lim \sup }}\,\left\| d\left( t \right)-{{d}_{f}} \right\|\le \zeta$, where $\zeta$ is a small nonnegative constant. The convergence of \eqref{solve_state_eq} is analyzed
\begin{equation}
\label{limit_state_eq}
\underset{t\to \infty }{\mathop{\lim }}\,\!{\bm{x}}\!\left( t \right)=\underset{t\to \infty }{\mathop{\lim }}\,\!\left( {{e}^{At}}{\bm{x}}\!\left( 0 \right)+\int_{0}^{t}{{{e}^{A\left( t-\tau  \right)}}}{\bm{B}d}\!\left( \tau  \right)d\tau  \right).
\end{equation}

First, we assume that \eqref{limit_state_eq} admits an equilibrium point, i.e., ${{x}_{2}}\!=\!0$ and ${x}_{1}$ converges to ${x}_{1f}$. Moreover, at steady state, the stable values of motion state ${\bm{x}}_{f}$ and input vector ${\bm{B}}_{f}$ are
\begin{equation}
\label{}
\bm{0}=A{{\bm{x}}_{f}}+{{\bm{B}}_{f}},{{\bm{x}}_{f}}={{\left( {{x}_{1f}},0 \right)}^\text{T}},{{\bm{B}}_{f}}={{\left( 0,\frac{k\cdot{{d}_{f}}}{m} \right)}^\text{T}},
\end{equation}
where, ${{\bm{x}}_{f}}=-{{A}^{-1}}{{\bm{B}}_{f}}$. At this point, we obtain ${{x}_{1f}}={{d}_{f}}$.
That is, if $A$ is invertible, the stable value ${x}_{1f}$ exists. It is then required to further verify that motion state ${x}_{1}$ converges asymptotically, with its limit necessarily equal to ${x}_{f}$. Accordingly, a new state is defined as
\begin{equation}
\label{}
\tilde{\bm{x}}\left( t \right)={\bm{x}}\left( t \right)-{{{\bm{x}}}_{f}}.
\end{equation}
Substituting into \eqref{state_eq}, we obtain
\begin{equation}
\label{}
\frac{d\tilde{\bm{x}}}{dt}=A{\tilde{\bm{x}}}\left( t \right)+{\bm{B}d(t)}-{{\bm{B}}_{f}}.
\end{equation}
By letting $\tilde{\bm{B}}\!\left( t \right)\!=\!\bm{B}d(t)-{{\bm{B}}_{f}}$, and following a procedure similar to \eqref{solve_state_eq}, the solution is given by
\begin{equation}
\label{}
{\tilde{\bm{x}}}\left( t \right)={{e}^{At}}{\tilde{\bm{x}}}\left( 0 \right)+\int_{0}^{t}{{{e}^{A\left( t-\tau  \right)}}}\tilde{\bm{B}}\left( \tau  \right)d\tau.
\end{equation}
Since the eigenvalues of $A$ have negative real parts, according to the Lyapunov stability criterion
\begin{equation}
\label{}
{{A}^\text{T}}M+MA=-Q,
\end{equation}
where $M$ and $Q$ are symmetric positive-definite matrices, and for a given $Q$, $M$ is unique. An energy function is
\begin{equation}
\label{}
V\left( {{t}} \right)={{\tilde{\bm{x}}}^\text{T}}M{\tilde{\bm{x}}}.
\end{equation}
Let $\dot{V}$ denote the time derivative of $V\!\left( {t} \right)$
\begin{equation}
\label{}
 \dot{V}=-{{\tilde{\bm{x}}}^\text{T}}Q{\tilde{\bm{x}}}+2{{\tilde{\bm{x}}}^\text{T}}M\tilde{\bm{B}}.
\end{equation}
According to the basic inequality:
\begin{equation}
\label{}
2{{\tilde{\bm{x}}}^\text{T}}M\tilde{\bm{b}}\le \varepsilon {{\left\| {\tilde{\bm{x}}} \right\|}^{2}}+\frac{1}{\varepsilon }{{\left\| M\tilde{\bm{B}} \right\|}^{2}},
\end{equation}
where $\varepsilon $ is an arbitrary positive scalar, $-{{\tilde{\bm{x}}}^\text{T}}Q{\tilde{\bm{x}}}\le -{{\lambda }_{\min }}\left( Q \right){{\left\| {\tilde{\bm{x}}} \right\|}^{2}}$, and ${{\lambda }_{\min }}\left( \cdot  \right)$ and ${{\lambda }_{\max }}\left( \cdot  \right)$ denote the minimum and maximum eigenvalues of the corresponding matrix, respectively. Hence, we have
\begin{equation}
\label{inequality_eq}
\dot{V}\le -{{\lambda }_{\min }}\left( Q \right){{\left\| {\tilde{\bm{x}}} \right\|}^{2}}+\varepsilon {{\left\| {\tilde{\bm{x}}} \right\|}^{2}}+\frac{1}{\varepsilon }{{\left\| M\tilde{\bm{B}} \right\|}^{2}}.
\end{equation}
Let $\varepsilon\!\!=\!\!\frac{1}{2}{{\lambda }_{\min }}\!\left( Q \right)$, and \eqref{inequality_eq} can be simplified as
\begin{equation}
\label{inequality2_eq}
 \dot{V}\le -\frac{1}{2}{{\lambda }_{\min }}\left( Q \right){{\left\| {\tilde{\bm{x}}} \right\|}^{2}}+\frac{2}{{{\lambda }_{\min }}\left( Q \right)}{{\left\| M\tilde{\bm{B}} \right\|}^{2}}.
\end{equation}

Moreover, since ${{\lambda }_{\min }}\!\left( M \right){{\left\| {\tilde{\bm{x}}} \right\|}^{2}}\!\le\!V\!\le\!{{\lambda }_{\max }}\!\left( M \right){{\left\| {\tilde{\bm{x}}} \right\|}^{2}}$, let ${l=2}/{{{\lambda }_{\min }}\!\left( Q \right)}\;$. By combining \eqref{inequality2_eq}, we obtain
\begin{equation}
\label{}
\frac{-{{\lambda }_{\min }}\left( Q \right)}{2{{\lambda }_{\max }}\left( M \right)}V+l{{\left\| M\tilde{\bm{B}} \right\|}^{2}}\ge \dot{V}.
\end{equation}

Let $W\!\!\left( \tau  \right)\!=\!{{e}^{r\tau }}V\!\!\left( \tau \right)$ and $r\!=\!{{{\lambda }_{\min }}\!\!\left( Q \right)}\!/\!{\left( 2{{\lambda }_{\max }}\!\left( M \right) \right)}\;$, thus
\begin{equation}
\label{inequality3_eq}
\dot{W}\!\left( \tau  \right)={{e}^{r\tau }}\left( rV\!\left( \tau \right)+\dot{V} \right)\le l{{e}^{r\tau }}{{\left\| M\tilde{\bm{B}} \right\|}^{2}},	
\end{equation}
where $\tau$ also denotes time variable. Integrating \eqref{inequality3_eq} yields
\begin{equation}
\label{inequality4_eq}
V\left( t \right)\le{{e}^{-rt}}{{\left. V \right|}_{\tau =0}} +\int_{0}^{t}{l{{e}^{-r\left( t-\tau  \right)}}}{{\left\| M\tilde{\bm{B}}\left( \tau  \right) \right\|}^{2}}d\tau.
\end{equation}
The limit of first term on the right-hand side of \eqref{inequality4_eq} is
\begin{equation}
\label{}
\underset{t\to \infty }{\mathop{\lim }}\,{{e}^{-rt}}{{\left. V \right|}_{\tau =0}}=0.
\end{equation}
For the second term, it is defined as follows

\begin{equation}
\label{}
\int_{0}^{t}{\varphi \left( \tau  \right)}d\tau\triangleq\int_{0}^{t}{l{{e}^{-r\left( t-\tau  \right)}}}{{\left\| M\tilde{\bm{B}}\left( \tau  \right) \right\|}^{2}}d\tau.
\end{equation}

since ${{e}^{-r\left(t-\tau\right)}}\!\le\!1$, and ${{\left\| M\tilde{\bm{B}}\!\left( \tau  \right) \right\|}^{2}}$ is bounded and converges to a neighborhood. For $\forall \varsigma\!\!>\!\!0$, there exists a time instant ${{T}_{0}}\!>\!0$ such that when $\tau\!>\!{{T}_{0}}$, ${{\left\| M\tilde{\bm{B}}\!\left( \tau  \right) \right\|}^{2}}\!\le\! \varsigma$. The term can be decomposed as
\begin{equation}
\label{}
\int_{0}^{t}{\varphi \left( \tau  \right)}d\tau\!\!=\!\!\int_{0}^{{T}_{0}}{\varphi \left( \tau  \right)}d\tau{+}\!\int_{{T}_{0}}^{t}{\varphi \left( \tau  \right)}d\tau\! .
\end{equation}
Over the interval $\left( 0,{{T}_{0}} \right)$, let  maximum of ${{\left\| M\tilde{\bm{B}}\left( \tau  \right) \right\|}^{2}}$ be $C$. Then, the first part can be written as
\begin{equation}
\label{}
\int_{0}^{{T}_{0}}{\varphi \left( \tau  \right)}d\tau\! \!\le \frac{lC{{e}^{-rt}}}{r}\left( {{e}^{r{{T}_{0}}}}-1 \right).
\end{equation}
Thus
\begin{equation}
\label{}
\underset{t\to \infty }{\mathop{\lim }}\,\int_{0}^{{T}_{0}}{\varphi \left( \tau  \right)}d\tau\!=0.
\end{equation}
For the second part
\begin{equation}
\label{}
\!\int_{{T}_{0}}^{t}{\varphi \left( \tau  \right)}d\tau\! \le \varsigma l \int_{{T}_{0}}^{t}{{{e}^{-r\left( t-\tau  \right)}}}d\tau \le \frac{\varsigma l }{r}.
\end{equation}
Therefore,
\begin{equation}
\label{}
\underset{t\to \infty }{\mathop{\lim \sup }}\,V\left( t \right)\le \frac{\varsigma l }{r}.
\end{equation}
Combining the condition ${{\lambda }_{\min }}\!\!\left( M \right)\!{{\left\| {\tilde{\bm{x}}} \right\|}^{2}}\le\!V$, it follows
\begin{equation}
\label{convergence_eq}
\underset{t\to \infty }{\mathop{\lim \sup }}\,\left\| \tilde{\bm{x}}\left( t \right) \right\|\le \mu \Rightarrow \underset{t\to \infty }{\mathop{\lim \sup }}\,\left\| \bm{x}\left( t \right)-{\bm{x}_{f}} \right\|\le \mu 
\end{equation}
where, $\mu$ is a small nonnegative constant.
Therefore, ${\bm{x}}\left( t \right)$ converges to $-{{A}^{-1}}{{\bm{B}}_{f}}={{\left( {{d}_{f}},0 \right)}^\text{T}}$, indicating that manipulator and target converge synchronously. 
In summary, the robot trajectory demonstrates target-constrained capability and meets the tracking requirement of eliminating steady-state errors. Nevertheless, to further regulate the trajectory—such as obstacle avoidance during tracking and grasping of moving targets—additional control is required to shape the trajectory. Therefore, the subsequent sections focus on the generation of dynamic constraint and the corresponding tracking control of manipulator.

\begin{figure}[!t]
\centering
\includegraphics[width=3.0in]{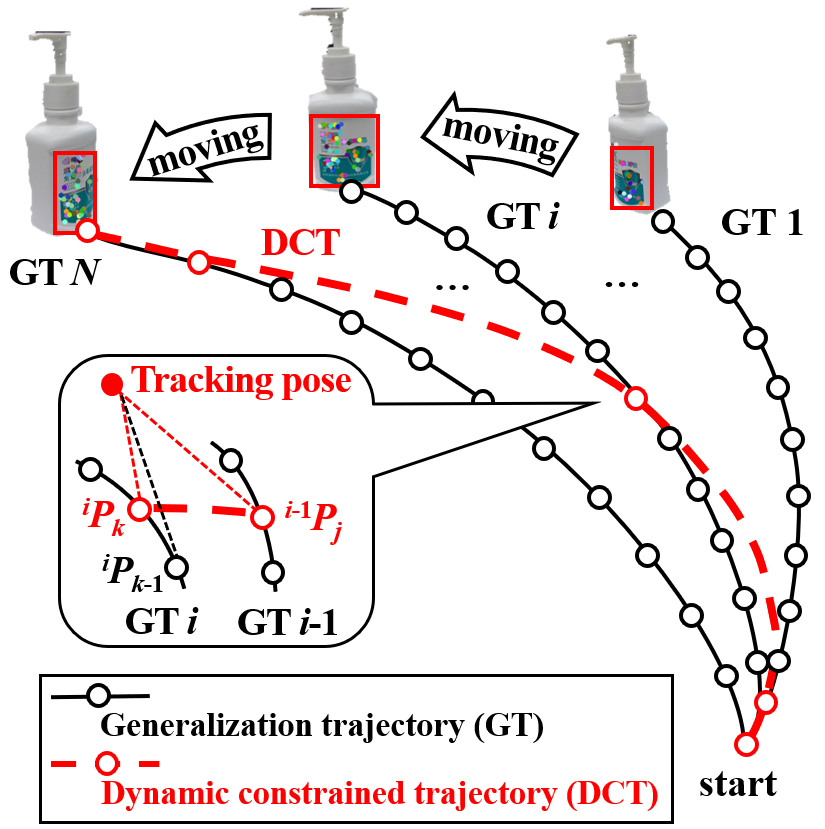}
\caption{Dynamic imitation-constrained trajectory generation}
\label{dynamic_constraint}
\end{figure}

\subsection{Trajectory Constraints for Tracking}
According to the perception prerequisite, it is assumed that real-time pose and feature points are known, implying that $d(t)$ is available, and \eqref{state_eq} can be simplified as:
\begin{equation}
\label{state_eq2}
\ddot{x}_{1}=\frac{c}{m} \left( \frac{k}{c} \left( d(t)-{x}_{1} \right)-\dot{x}_{1} \right).
\end{equation}

Given the smoothness and local support properties of Gaussian functions, a trajectory modulation term $f$ composed of Gaussian basis functions ${{\psi }_{i}}$ is introduced into \eqref{state_eq2}, as
\begin{equation}
\label{state_control_eq}
\left\{ \begin{aligned}
  & \ddot{x}_{1}=\frac{c}{m} \left( \frac{k}{c} \left( d-{x}_{1} \right)-\dot{x}_{1} \right)+f  \\
  & f=\frac{\sum\nolimits_{i=1}^{N}{{{\psi }_{i}}{{\omega }_{i}}}}{\sum\nolimits_{i=1}^{N}{{{\psi }_{i}}}}s(t)\left( d-{{x}_{0}} \right)  \\
\end{aligned} \right.,
\end{equation}
where, ${{\omega }_{i}}$ denotes the weight of the ${{\psi }_{i}}$, ${x}_{0}$ is the starting point of trajectory, and $\dot{s}\left( t \right)=-\gamma s\left( t \right)$ is the phase function, decreasing monotonically from 1 to 0, where $\gamma$ is a convergence factor regulating the convergence speed, and it is typically set to 1; therefore, $f$ does not affect the convergence property of \eqref{convergence_eq}. However, it suffers from the following limitations:
\begin{itemize}
\item It is primarily designed for imitation operation involving static targets.
\item It cannot reproduce trajectories with identical start and end points but different intermediate paths, due to the $(d-{{x}_{0}})$ in $f$ which vanishes when $d={x}_{0}$. 
\item Generalization requires variable convergence times. 
\end{itemize}

Therefore, we first introduce a temporal scaling transformation. Given a desired positive scaling factor $\eta$, the transformed time ${{t}_{c}}$ and original time ${{t}_{p}}$ are related by:
\begin{equation}
\label{time_trans_eq}
{{t}_{c}}=\eta {{t}_{p}}
\end{equation}
Thus, the relationship between the new and the original state is given as
\begin{equation}
\label{new_state}
\left\{ \begin{array}{*{35}{l}}
   {{x}_{1p}}\left( {{t}_{p}}  \right)={{x}_{1c}}\left( {{t}_{c}} \right)={{x}_{1c}}\left( \eta {{t}_{p}} \right)  \\
   {{{\dot{x}}}_{1p}}\left( {{t}_{p}} \right)=\eta {{{\dot{x}}}_{1c}}(\eta {{t}_{p}})=\eta {{{\dot{x}}}_{1c}}\left( {{t}_{c}} \right)  \\
   {{{\ddot{x}}}_{1p}}\left( {{t}_{p}} \right)={{\eta }^{2}}{{{\ddot{x}}}_{1c}}(\eta {{t}_{p}})={{\eta }^{2}}{{{\ddot{x}}}_{1c}}\left( {{t}_{c}} \right)  \\
\end{array} \right..
\end{equation}
Similarly, \eqref{time_trans_eq} is incorporated into the $s(t)$ and ${\psi}_{i}(t)$, as
\begin{equation}
\label{new_state}
\left\{ \begin{aligned}
  & {{s}_{c}}\left( {{t}_{c}} \right)=\exp \left( -\frac{\gamma {{t}_{c}}}{\eta } \right) \\ 
 & {{\psi }_{ci}}\left( {{t}_{c}} \right)=\exp \left( -\frac{1}{2\sigma _{i}^{2}}{{\left( {{s}_{c}}-{{c}_{i}} \right)}^{2}} \right) \\ 
\end{aligned}\right.
\end{equation}
Meanwhile, to accommodate learning trajectories with identical start and end points, we decouple the shape and amplitude components in the $f$, yielding
\begin{equation}
\label{new_f}
{{f}_{c}}=\left( \frac{\sum\nolimits_{i=1}^{N}{{{\psi }_{ci}}{{\omega }_{i}}}}{\sum\nolimits_{i=1}^{N}{{{\psi}_{ci}}}}+\left( d-{{x}_{0}} \right) \right){{s}_{c}}.
\end{equation}
By substituting  \eqref{new_state} and  \eqref{new_f} into \eqref{state_control_eq}, we obtain
\begin{equation}
\label{m_state_control_eq}
{{\eta }^{2}}{{\ddot{x}}_{1c}}=\frac{c}{m} \left( \frac{k}{c} \left( d-{{x}_{1c}} \right)-\eta {{{\dot{x}}}_{1c}} \right)+{{f}_{c}}.
\end{equation}
Then we substitute demonstration trajectory into \eqref{m_state_control_eq} to derive Gaussian fitting term $G({{t}_{c}})$ as
\begin{equation}
\label{gauss_fit_eq}
\begin{aligned}
  & {{G}_\text{demo}}\left( {{t}_{c}} \right)\triangleq \frac{\sum\nolimits_{i=1}^{N}{{{\psi }_{ci}}{{\omega }_{i}}}}{\sum\nolimits_{i=1}^{N}{{{\psi }_{ci}}}}{{s}_{c}} \\ 
 & {{t}_{c}}=j\cdot {{t}_\text{step}}\text{, }j=1,2,\ldots ,n \\ 
\end{aligned},
\end{equation}
where, $n$ denotes the total number of time steps, and ${t}_\text{step}$ is the average time interval between consecutive points in the demonstration trajectory. By rearranging \eqref{gauss_fit_eq}, we obtain
\begin{equation}
\label{}
\sum\limits_{i=1}^{N}{{{\psi }_{ci}}\left( {G(t)}-{{\omega }_{i}}{{s}_{c}} \right)=0}.
\end{equation}
For each ${{\omega }_{i}}$, Locally Weighted Regression (LWR) method is employed. Specifically, loss function is constructed as
\begin{equation}
\label{deqn_ex1a}
L\left( {{\omega }_{i}} \right)=\sum\limits_{t={t}_{1}}^{{{t}_{k}}}{{{\psi }_{ci}}{{\left( {G(t)}-{{\omega }_{i}}{{s}_{c}} \right)}^{2}}}.
\end{equation}
Then, by $\frac{\partial L\left( {{\omega }_{i}} \right)}{\partial {{\omega }_{i}}}=0$, ${{\omega }_{i}}$ can be computed as 
\begin{equation}
\label{deqn_ex1a}
{{\omega }_{i}}=\frac{{{\bm{S}}^{T}}{{\mathit{\Psi} }_{ci}}}{{{\bm{S}}^{T}}{{\mathit{\Psi} }_{ci}}\bm{S}}{{\bm{G}}_{e}},
\end{equation}
\begin{equation}
\label{deqn_ex1a}
\bm{S}={{\left( {{s}_{c}}\left( {t}_{1} \right),{{s}_{c}}\left( {t}_{2} \right),...,{{s}_{c}}\left( {t}_{n} \right) \right)}^\text{T}},
\end{equation}
\begin{equation}
\label{deqn_ex1a}
{{\mathit{\Psi} }_{ci}}=I\cdot {{\left( {{\psi }_{ci}}\left( {t}_{1} \right),{{\psi }_{ci}}\left( {t}_{2} \right),...,{{\psi }_{ci}}\left( {t}_{n} \right) \right)}^\text{T}},
\end{equation}
\begin{equation}
\label{deqn_ex1a}
{{\bm{G}}_{e}}={{\left( G\left( {t}_{1} \right),G\left( {t}_{2} \right),...,G\left( {t}_{n} \right) \right)}^\text{T}}.
\end{equation}

\begin{figure}[!t]
\centering
\includegraphics[width=3.5in]{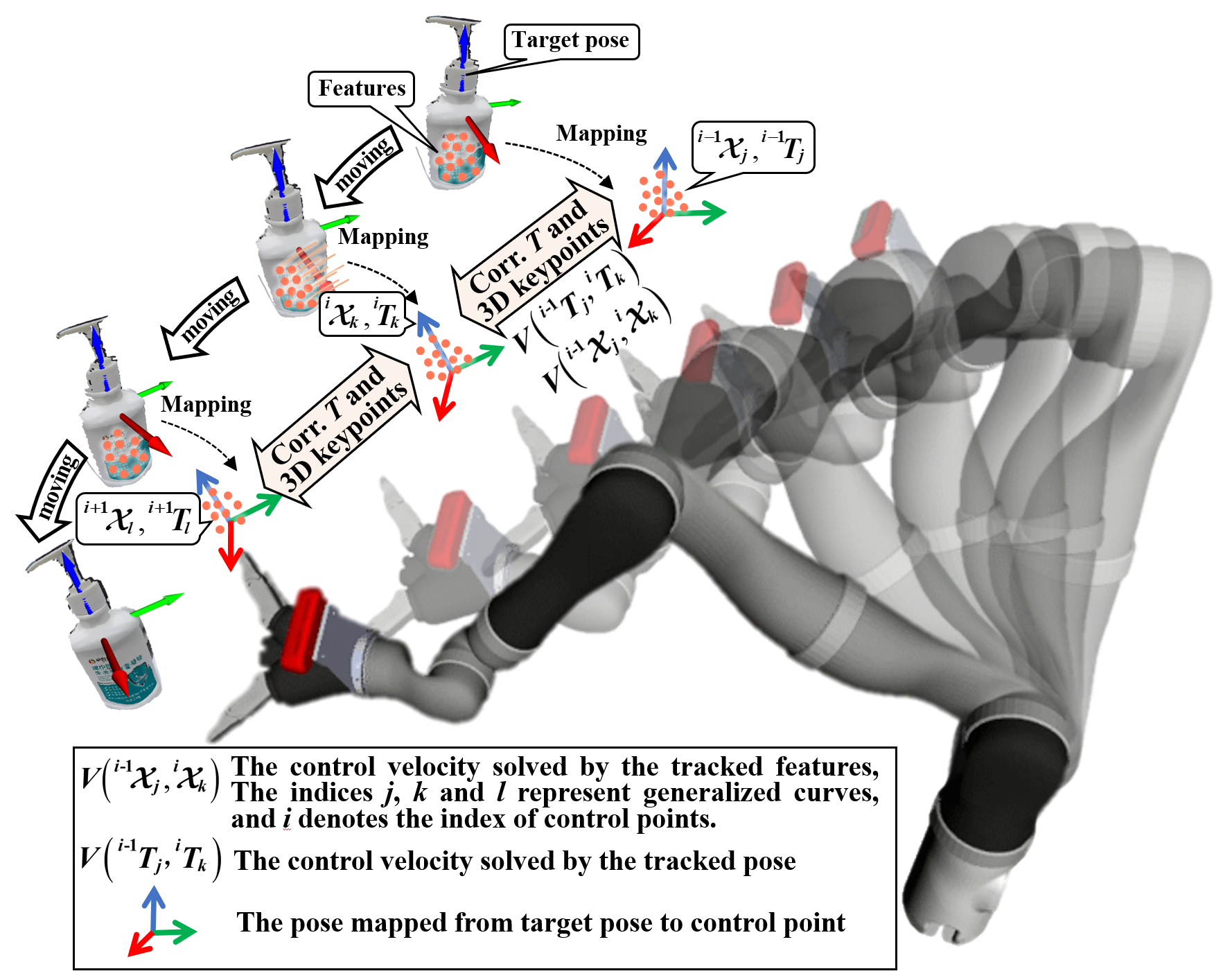}
\caption{Mapping the target pose and texture features to the  control velocities of trajectory (Corr.: Corresponding)}
\label{velocity_cotrol_fig}
\end{figure}

Finally, by substituting ${{\omega }_{i}}$ into \eqref{m_state_control_eq}, generalization model is obtained. And the system adaptively determines constraint points from a sequence of real-time generalized trajectories (GT), constructing a dynamic constrained trajectory (DCT) as shown in Fig.~\ref{dynamic_constraint}.
Current motion point ${}^{i-1}{{\bm{P}}_{j}}$ is the $j$-th control point. Under the guidance of moving targets, next generalized trajectory is denoted as $i$, from which the $k$-th control point is determined according to the convergence criterion. Accordingly, the next control point ${}^{i}{{\bm{P}}_{k}}$, its neighboring point ${}^{i}{{\bm{P}}_{k-1}}$, and the current control point ${}^{i-1}{{\bm{P}}_{j}}$ must satisfy the following condition:
\begin{equation}
\label{dynamic_condition}
{{\left\| {}^{i}{{\bm{P}}_{k}}-{{\bm{P}}_{g}} \right\|}_{2}}<{{\left\| {}^{i-1}{{\bm{P}}_{j}}-{{\bm{P}}_{g}} \right\|}_{2}}<{{\left\| {}^{i}{{\bm{P}}_{k-1}}-{{\bm{P}}_{g}} \right\|}_{2}},
\end{equation}
where, ${{\bm{P}}_{g}}$ is the position of desired tracking pose of next generalized trajectory. To enable efficient determination of control points, following strategy is adopted. Since adjacent generalized trajectories $i-1$ and $i$ are similar in shape and distribution, we initialize the candidate index by $k = j$, that is, the $j$-th point on $i$-th trajectory is first taken as next candidate control point.
Then, the \eqref{dynamic_condition} is evaluated, and the candidate is updated toward the preceding or succeeding point until the criterion is satisfied.
By iteratively applying this mechanism, a sequence of control points is constructed to form a dynamic control path.

\begin{figure*}[htbp]
\centering
\includegraphics[width=7.0in]{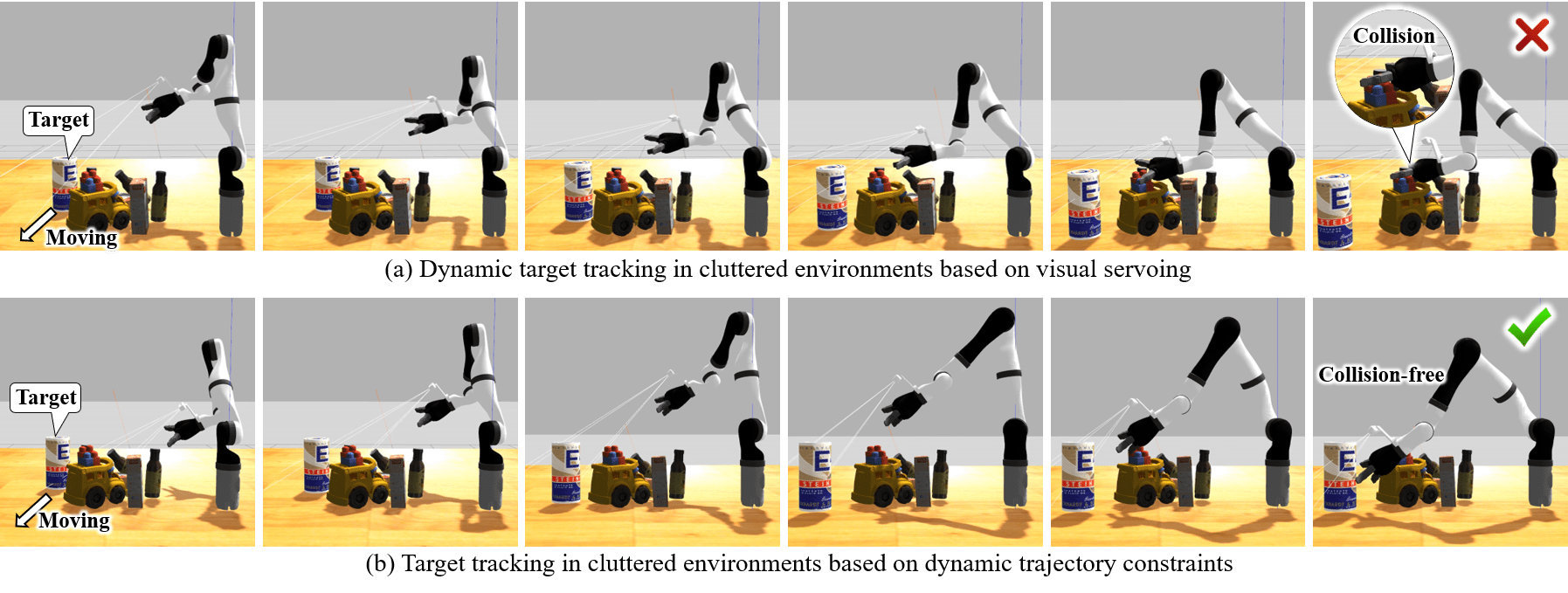}
\caption{Ablation experiments for moving targets tracking}
\label{simulation}
\end{figure*}

Importantly, the DCT provides an explicit intermediate motion constraint rather than a pure instantaneous velocity mapping. Specifically, the constraint points enforce a monotonic decrease of the distance-to-goal, which guarantees continuous progress toward the target and prevents back-and-forth oscillations caused by noisy feature residuals. Moreover, since each determined point ${}^{i}{{\bm{P}}_{k}}$ is chosen from a locally smooth generalized trajectory and constrained by its neighboring point ${}^{i}{{\bm{P}}_{k-1}}$, the resulting DCT inherits local continuity from the trajectory generator, yielding a smooth, well-shaped motion that can further accommodate task-level requirements (e.g., obstacle avoidance) during tracking.

\subsection{Control of Robotic Tracking}
Although the convergence and constraint are presented for one representative motion dimension, the same formulation is applied to all task space of end-effector. The translational and rotational components are then assembled into six-dimensional spatial velocity, so that the dynamic constrained trajectory can be tracked through pose differentials or multiple feature correspondences as in Fig.~\ref{velocity_cotrol_fig}. Given the short time interval between control points along the dynamic trajectory, the variation between current grasp pose and next control point can be treated as a small transformation.
The target frames and texture points of the next and current generalized trajectories are mapped to their respective control point, resulting in (${{T}_{next}}$, ${\mathcal{X}}_{next}$) and (${{T}_{curr}}$, ${\mathcal{X}}_{curr}$).
For visual methods that only perform target pose tracking, the differential pose transformation is used as
\begin{equation}
\label{deqn_ex1a}
{{T}_{next}}=\text{Trans}(dx,dy,dz)\text{Rot}(\delta x,\delta y,\delta z){{T}_{curr}},
\end{equation}
where $\text{Trans}(dx,dy,dz)$ denotes the differential translation, and $\text{Rot}(\delta x,\delta y,\delta z)$ represents the differential rotation.
With the small motion assumption, the higher-order components induced by the rotation order can be neglected due to their minimal influence. Thus, transformation can be described as 
\begin{equation}
\label{deqn_ex1a}
dT={{T}_{next}}-{{T}_{curr}}=\left[ \Delta  \right]{{T}_{curr}},
\end{equation}
\begin{equation}
\label{deqn_ex1a}
\left[ \Delta  \right]\approx \left( \begin{matrix}
   0 & -\delta z & \delta y & dx  \\
   \delta z & 0 & -\delta x & dy  \\
   -\delta y & \delta x & 0 & dz  \\
   0 & 0 & 0 & 0  \\
\end{matrix} \right).
\end{equation}

Based on the above, the motion ${{\left[ \Delta  \right]}^{\vee }}$ between adjacent poses can be approximated instantaneous motion, and $\vee $ denotes the operator that extracts the motion vector ${{\left( dx,dy,dz,\delta x,\delta y,\delta z \right)}^\text{T}}$. Finally, end-effector velocity ${{\bm{V}}_{b}}={{({{\bm{v} }^\text{T}},{{\bm{\omega} }^\text{T}})}^\text{T}}$ is acquired according desired time $\Delta t$ specified by application, as ${{\bm{V}}_{b}}={{{\left[ \Delta  \right]}^{\vee }}}/{\Delta t}\;$. 

For visual methods that track target keypoints, the control velocity can be obtained by the following formulation.
Specifically, the corresponding feature points ${\bm{p}_{next}}$ and ${\bm{p}_{curr}}$ are selected from the ${\mathcal{X}}_{next}$ and ${\mathcal{X}}_{curr}$, and the feature velocity ${{\dot{\bm{p}}}_{f}}$ can be computed based on the desired time $\Delta t$.
Therefore, the linear and angular velocities ${{\bm{V}}_{b}}$ of the end-effector are related to the feature point velocity as
\begin{equation}
\label{deqn_ex1a}
{{\dot{\bm{p}}}_{f}}=\bm{\omega} \times {{\bm{p}}_{f}}+\bm{v} =\left( \begin{matrix}
   {{{\dot{x}}}_{f}}  \\
   {{{\dot{y}}}_{f}}  \\
   {{{\dot{z}}}_{f}}  \\
\end{matrix} \right)=\left( \begin{matrix}
   {{z}_{f}}{{\omega }_{y}}-{{y}_{f}}{{\omega }_{z}}+{{v}_{x}}  \\
   {{x}_{f}}{{\omega }_{z}}-{{z}_{f}}{{\omega }_{x}}+{{v }_{y}}  \\
   {{y}_{_{f}}}{{\omega }_{x}}-{{x}_{f}}{{\omega }_{y}}+{{v }_{z}}  \\
\end{matrix} \right).
\end{equation}

By aggregating $n$ mapped keypoint pairs, we construct following overdetermined system of equations
\begin{equation}
\label{}
\begin{aligned}
  & \left\{ \begin{aligned}
  & {{{\dot{\bm{p}}}}_{f1}}={{J}_{f1}}{{\bm{V}}_{b}}  \\ 
 & \text{     }\cdot \cdot \cdot  \\ 
 & {{{\dot{\bm{p}}}}_{fn}}={{J}_{fn}}{{\bm{V}}_{b}} \\ 
\end{aligned} \right.\to {{{\dot{\bm{P}}}}_{f}}={{L}_{f}}{{\bm{V}}_{b}} \\ 
 & {{{\dot{\bm{P}}}}_{f}}\in {{\mathbb{R}}_{3n\times 1}}\text{, }{{L}_{f}}\in {{\mathbb{R}}_{3n\times 6}}\text{, }{{\bm{V}}_{b}}\in {{\mathbb{R}}_{6\times 1}} \\ 
\end{aligned},
\end{equation}
\begin{equation}
\label{deqn_ex1a}
{{J}_{fi}}=\left( \begin{array}{*{35}{l}}
   1 & 0 & 0 & 0 & {{z}_{{{fi}}}} & -{{y}_{fi}}  \\
   0 & 1 & 0 & -{{z}_{fi}} & 0 & {{x}_{fi}}  \\
   0 & 0 & 1 & {{y}_{fi}} & -{{x}_{fi}} & 0  \\
\end{array} \right).
\end{equation}
The end-effector velocity is then given by
\begin{equation}
\label{deqn_ex1a}
{{\bm{V}}_{b}}={{\left( L_{f}^{T}{{L}_{f}} \right)}^{-1}}{L_{f}^\text{T}}{{\dot{\bm{P}}}_{f}},
\end{equation}

Next, the ${{\bm{V}}_{b}}$ is transformed into the velocity with respect to gripper frame, and is then represented in the base frame, as ${\bm{V}}_{s}$. Finally, ${\bm{V}}_{s}$ is in conjunction with spatial Jacobian matrix ${J}$ to compute joint velocity ${\bm{V}}_{j}$,
\begin{equation}
\label{deqn_ex1a}
{{\bm{V}}_{s}}=\left( \begin{matrix}
   {{R}_{curr}} & 0  \\
   0 & {{R}_{curr}}  \\
\end{matrix} \right)\text{Ad}_\text{V}^{-1}\left( {{T}_{curr}} \right){{\bm{V}}_{b}},
\end{equation}
\begin{equation}
\label{}
\text{Ad}_{\text{V}}^{-1}\left( {{T}_{curr}} \right)=\left( \begin{matrix}
   R_{curr}^\text{T} & -R_{curr}^\text{T}{{\left[ {\bm{p}_{curr}} \right]}_{\times }}  \\
   0 & R_{curr}^\text{T}  \\
\end{matrix} \right),
\end{equation}
\begin{equation}
\label{}
\text{ }{{\left[ {\bm{p}_{curr}} \right]}_{\times }}=\left( \begin{matrix}
   0 & -{{p}_{curr,z}} & {{p}_{curr,y}}  \\
   {{p}_{curr,z}} & 0 & -{{p}_{curr,x}}  \\
   -{{p}_{curr,y}} & {{p}_{curr,x}} & 0  \\
\end{matrix} \right),
\end{equation}
\begin{equation}
\label{deqn_ex1a}
{{\bm{V}}_{j}}={{{\left( {{J}^\text{T}}J \right)}^{-1}}{{J}^\text{T}}}{{\bm{V}}_{s}},
\end{equation}
where ${{R}_{curr}}$ and ${\bm{p}_{curr}}={{\left( {{p}_{curr,x}},{{p}_{curr,y}},{{p}_{curr,z}} \right)}^\text{T}}$ denote the rotational and translational components of ${{T}_{curr}}$.
\begin{figure*}[htbp]
\centering
\includegraphics[width=7.0in]{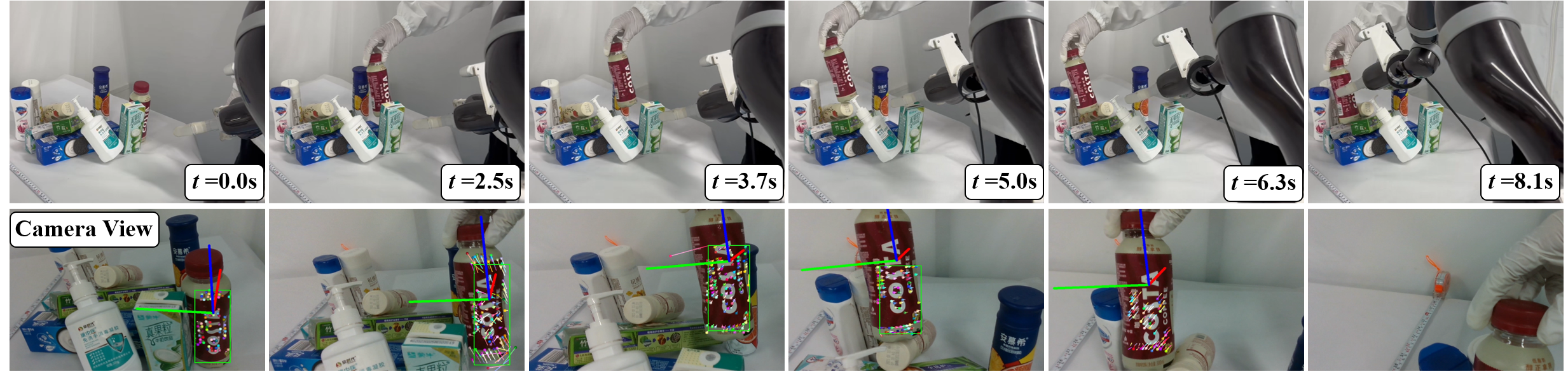}
\caption{Tracking process for mixed translational and rotational disturbances of targets}
\label{dynamic_6D_tracking}
\end{figure*}

\section{EXPERIMENTS}
This section presents experimental comparisons between the proposed dynamic tracking method and other state-of-the-art technology. The demonstration trajectories are collected under static conditions, including motion patterns such as approaching target and avoiding obstacles.
First, ablation experiments are conducted to evaluate target tracking in complex environments. The feature-guided visual servoing method imposes weak constraints on the intermediate trajectory and directly maps the residuals of key features to control velocities. As a result, the manipulator tends to collide with surrounding objects. In contrast, the constrained servo method first learns an obstacle-avoidance trajectory from demonstrations, and subsequently generates a dynamic constraint according to the target motion, thereby achieving safe and collision-free tracking as in Fig.~\ref{simulation}.

In real-world comparative experiments, the primary hardware used includes a Kinova robotic arm (with a maximum joint velocity set to ${\pi }/{3}\;$), a RealSense depth camera.
For a fair comparison, all methods are executed under the same initial configurations, target-motion scripts, and control interface. The robotic controller runs at 100Hz for all methods, and each baseline is tuned following its original paper (or released implementation) with parameters fixed across all trials. The experiments cover 20 diverse scenarios with varying target motions (translation, rotation, and mixed disturbances), illumination levels, and background conditions. To eliminate bias from different disturbance realizations, we replay identical target disturbances for every method and repeat each setting for the same number of attempts. A trial is counted as successful if the end-effector reaches the graspable pose region and completes the grasp without collision or target loss; otherwise it is marked as failure. Then we compare several mainstream tracking methods. The accuracy is evaluated by measuring the error between the expected and actual poses of the wrist-mounted camera mounted on arm. The expected camera pose is determined by these tracking methods, while the actual pose is obtained from hardware feedback when robot executes. The position error is decomposed into three translational degrees of freedom. For orientation error, we adopt evaluation metrics commonly used in SLAM for trajectory error. Specifically, the Absolute Pose Error (APE) is employed to quantify the rotational deviation. The above error metrics are defined as follows
\begin{equation}
\label{deqn_ex1a}
\left\{ \begin{matrix}
   {{e}_{x,t}}={{x}_{e,t}}-{{x}_{t}}  \\
   {{e}_{y,t}}={{y}_{e,t}}-{{y}_{t}}  \\
   {{e}_{z,t}}={{z}_{e,t}}-{{z}_{t}}  \\
\end{matrix} \right.,
\end{equation}
\begin{equation}
\label{deqn_ex1a}
{{E}_{t}}=R_{e,t\,}^{-1}{{R}_{t}},\text{ }AP{{E}_{t}}=\left\| {{E}_{t}}-{{I}_{3\times 3}} \right\|,
\end{equation}
where, ${{e}_{x,t}}$, ${{e}_{y,t}}$, and ${{e}_{z,t}}$ represent the position errors along the $x$-, $y$-, and $z$-axes at time step $t$, respectively, while $\text{APE}_t$ denotes the orientation error. The convergence behavior is characterized by metrics such as settling time, and the presence of overshoot in the trajectory (Based on the specification of employed arm, control error is approximately 3.9 mm. Considering the tolerance for reliable grasping of target ($\pm$10 mm), fluctuations exceeding 10 mm are treated as oscillation and overshoot induced by relative motion during tracking.).

\begin{figure*}[htbp]
\centering
\includegraphics[width=7.0in]{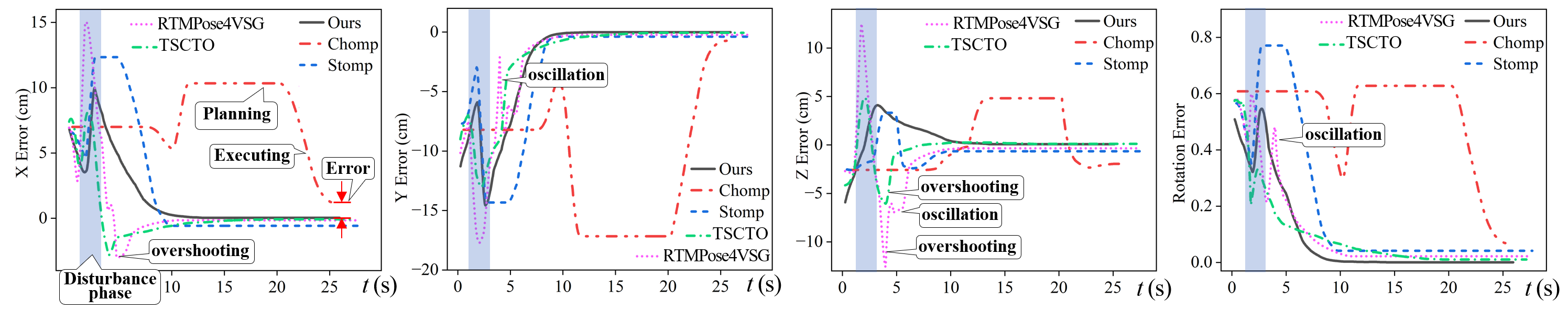}
\caption{Tracking errors for mixed translational and rotational disturbances of targets}
\label{6D_tracking_error}
\end{figure*}

\begin{figure}[!t]
\centering
\includegraphics[width=3.5in]{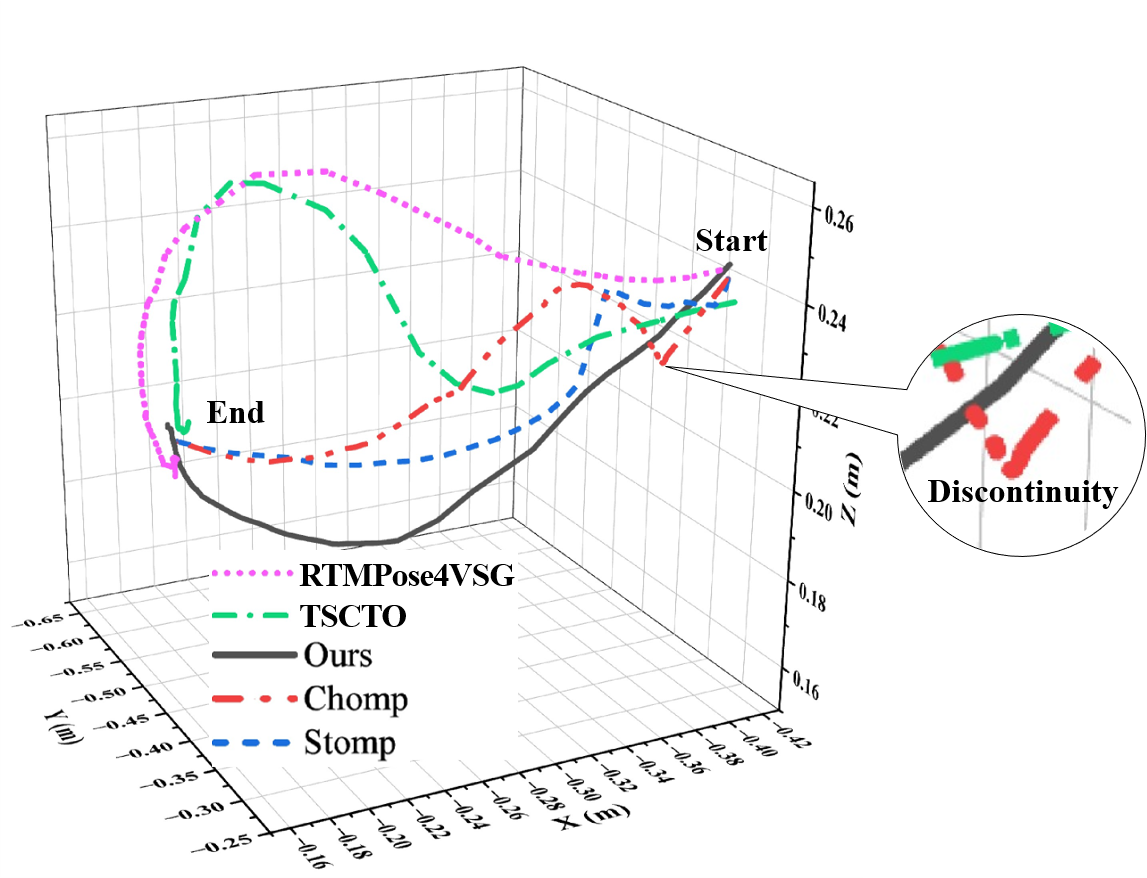}
\caption{Real trajectory curves for tracking moving target}
\label{tracking_traj}
\end{figure}

In dynamic scenarios, a combination of translational and rotational disturbances is applied to target within initial 0-5s as in Fig.~\ref{dynamic_6D_tracking}. Table~\ref{tracking_table} and Fig.~\ref{6D_tracking_error} report the convergence time, convergence accuracy, and success rate of the tracking task. 
Fig.~\ref{tracking_traj} further illustrates the specific convergence trajectories under the mixed disturbances shown in Fig.~\ref{dynamic_6D_tracking}.

\begin{table}[!t]
\centering
\setlength{\tabcolsep}{0.5mm}{
\begin{tabular}{cccccccc}
\toprule
Method & Cat. & Att. & Scen. & Dur. (s) & PE (mm) & SR (\%) & OS/OC \\
\midrule
RTMPose4VSG \cite{ref_Luo} &S & 150 &20 & 6.4 ± 1.0 & 8.0 ± 2.1 & 83 & 2 \\
TSCTO \cite{ref_shao} & S & 150 &20  & 8.3 ± 1.1 & 6.0 ± 1.8 & 79 & 1 \\
OMPL-CHOMP \cite{ref_OMPL} & G & 150 &20 & 23.4 ± 0.9 & 14 ± 3.7 & 80 & - \\
OMPL-STOMP \cite{ref_OMPL} & G & 150 &20 & 10.1 ± 0.9 & 18 ± 4.9 & 75 & - \\
\textbf{Ours} & \textbf{S} & \textbf{150} & \textbf{20} & \textbf{7.0 ± 0.8} & \textbf{4.0 ± 1.2} & \textbf{87} & \textbf{0}\\
\bottomrule
\end{tabular}}
\caption{Grasping experiment results for moving targets (Cat.: Category, Att.: Attempts, PE:Pose Error, SR: Success Rate, OS/OC: Overshoot/Oscillation, S: Servoing-based; G: Global planning-based)}
\label{tracking_table}
\end{table}
In the comparative experiments, our method distributes target's motion across all control points in the form of imitation trajectory constraints. As a result, the entire trajectory remains coherent as in Fig.~\ref{tracking_traj}, with consistent convergence in all motion dimensions as in Fig.~\ref{6D_tracking_error}. Moreover, proposed methods achieve real-time response to moving targets and high tracking accuracy (4 mm) in  Fig.~\ref{6D_tracking_error} and  Table~\ref{tracking_table}.

In the remaining methods, the RTMPose-based visual servoing (RTMPose4VSG) \cite{ref_Luo} achieves fast response but lacks effective constraint mechanisms due to the direct mapping from 2D feature variations to 6D spatial velocities. During tracking, motion disturbances cause keypoint fluctuations that are propagated through the image Jacobian matrix, leading to oscillations in the tracking velocity.
As for time-space trajectory optimization (TSCTO), a large number of non-differentiable trajectory segments are generated, introducing estimation errors in the computation of the state parameters. In addition, the robot’s predictive planning of local motion lags behind target's motion, resulting in overshoot as in Fig.~\ref{tracking_traj} and Table~\ref{tracking_table}. 
And in Fig.~\ref{6D_tracking_error}, the OMPL-based servo methods (CHOMP and STOMP) exhibit slow response to moving targets. And the tracking process separates into two phases-planning and execution-with long planning duration, and some errors exceed 10 mm. Meanwhile, the trajectory does not exhibit a smooth transition as in Fig.~\ref{tracking_traj}.

\section{CONCLUSIONS}

In this paper, we present a trajectory-constrained tracking method for moving targets, to improve motion continuity, interpretability, and task-level controllability. Specifically, a convergence model for manipulator's motion toward moving targets is established and analyzed by treating target motion as a time-varying input. 
Based on this model, a dynamic imitation-constrained mechanism is developed through time-scale deformation and trajectory modulation with decoupled shape and amplitude components. 
This generates a series of trajectories in real time and adaptively determines trajectory points to construct dynamic constraints. 
Then, a trajectory tracking controller is designed, where the robot’s spatial velocity is optimized by real-time target pose or key feature motion. The results of simulation and real-world experiments indicate that distributing target's motion across a sequence of constrained control points is beneficial for suppressing oscillation, reducing overshoot, and improving adaptability to irregular target motion.

\end{document}